\documentclass[10pt,twocolumn,letterpaper]{article}

\usepackage[pagenumbers]{configs/cvpr} %

\usepackage[pagebackref,breaklinks,colorlinks,allcolors=cvprblue]{hyperref}

\usepackage{xspace}        %
\usepackage{fontawesome}
\usepackage{lipsum}
\usepackage[normalem]{ulem}
\usepackage{ifthen}
\usepackage{xstring}
\usepackage{xcolor, colortbl}
\usepackage{multirow}
\usepackage{subcaption}
\usepackage{booktabs}
\usepackage[symbol]{footmisc}
\usepackage{stfloats}
\usepackage{balance}
\usepackage{float}
\usepackage{pifont}

\usepackage{siunitx}
\usepackage[capitalize,noabbrev]{cleveref}
\usepackage{placeins}

\newcommand{\qheading}[1]{\noindent\textbf{#1.}}

\definecolor{cvprblue}{rgb}{0.21,0.49,0.74}
\definecolor{salmonpink}{rgb}{1, 0.5, 0.4}
\definecolor{turquoise}{cmyk}{0.65,0,0.1,0.3}
\definecolor{purple}{rgb}{0.65,0,0.65}
\definecolor{dark_green}{rgb}{0, 0.5, 0}
\definecolor{orange}{rgb}{0.8, 0.6, 0.2}
\definecolor{darkred}{rgb}{0.6, 0.1, 0.05}
\definecolor{blueish}{rgb}{0.0, 0.3, .6}
\definecolor{light_gray}{rgb}{0.7, 0.7, .7}
\definecolor{pink}{rgb}{1, 0, 1}
\definecolor{greyblue}{rgb}{0.25, 0.25, 1}
\definecolor{forestgreen}{rgb}{0.0, 0.2, 0.13}
\definecolor{darkolivegreen}{rgb}{0.33, 0.42, 0.18}
\definecolor{teaserred}{rgb}{1, 0.588, 0.553}
\definecolor{teaserblue}{rgb}{0.337, 0.757, 1}

\hypersetup{
  colorlinks=true,
  allcolors=cvprblue,
  breaklinks=true
}

\providecommand{\highlightCOLOR}[1]{#1}

\newcommand{\moniker}{NGL}
\newcommand{\methodname}{\mbox{\highlightCOLOR{\moniker}}\xspace}

\newcommand{\dsl}{NGL}

\newcommand{\asosdatasetname}{ASOS}

\newcommand{\asoslabeled}{\emph{ASOS\_labeled}}

\usepackage{algorithm}
\usepackage{algpseudocode}

\usepackage[frozencache=true,cachedir=minted-cache]{minted}
\usepackage{multicol}

\def\paperID{xxx} %
\def\confName{CVPR}
\def\confYear{2026}

\title{\methodname: Natural Garment Language for Training-Free Sewing Pattern Estimation}

\author{Anna Badalyan$^{1}$ \quad Pratheba Selvaraju$^{1}$ \quad Giorgio Becherini$^{1}$ \quad Omid Taheri$^{1}$ \\ \quad Victoria Fernández Abrevaya$^{1}$ \quad Michael Black$^{1}$ \\
\small{$^{1}$Max Planck Institute for Intelligent Systems}
}

\usepackage{xspace}

\begin{document}
 \twocolumn[{%
   \renewcommand\twocolumn[1][]{#1}%
  \maketitle
   \vspace*{-1.2cm}
    \begin{center}
     \centerline{ \includegraphics[trim=0 0mm 0mm 0mm, clip=true, width=1.0 \linewidth]{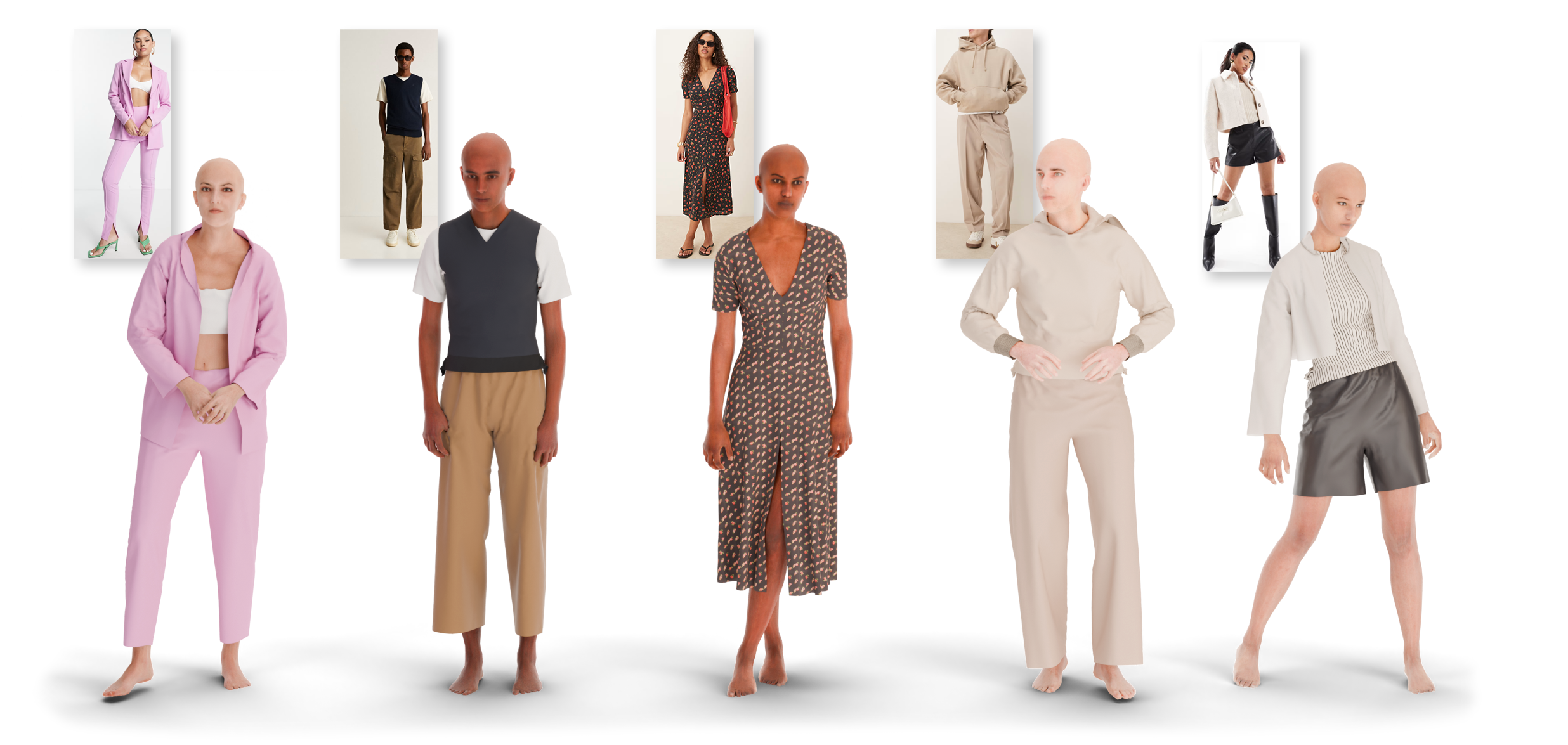}}
  \vspace*{-2.6em}
   \end{center}
   \begin{center}
 \captionof{figure}{
 \qheading{3D garment reconstruction by \moniker{}} 
 Given an image of a clothed person, our method estimates sewing patterns in a training-free manner, handling both single and multi-layer outfits.
 }
 \label{fig:teaser}
 \end{center}%
}]

\begin{abstract}
Estimating sewing patterns from images is a practical approach for creating high-quality 3D garments, but it remains challenging due to the scarcity of paired real-world image and sewing-pattern data.
Existing methods address this limitation by training vision-language models (VLMs) to learn low-level sewing-pattern representations from synthetic garments sampled from parametric garment models. 
However, they often struggle to generalize to in-the-wild images, fail to capture real-world correlations between garment parts, and are %
restricted to single-layer outfits.
In contrast, we observe that VLMs are effective at describing garments in natural language, %
but mapping these descriptions into valid sewing patterns remains difficult.
To bridge this gap, we propose \dsl{} (Natural Garment Language), a novel domain-specific language that represents garments in terms %
aligned with VLMs’ natural descriptive abilities. 
Leveraging \dsl, we introduce a fully training-free pipeline that queries large VLMs to extract structured garment specifications and deterministically converts them into valid sewing patterns.
We evaluate our method on the Dress4D, CloSe and a newly collected dataset of 253 in-the-wild fashion images. Our approach achieves state-of-the-art performance on standard geometry metrics and is preferred
in both human and GPT-based perceptual evaluations compared to existing baselines.
Furthermore, \moniker\/ recovers multi-layer outfits whereas competing methods focus mostly on single-layer garments, highlighting its strong generalization to real-world images even with occluded parts.
These results demonstrate that an efficient garment representation is critical for sewing pattern estimation with VLMs. Our code and data will be released for research use.
\end{abstract}

\maketitle

\section{Introduction}
Digital garments are used in many applications such as animation and games, virtual try-on, AR/VR telepresence, and automated fashion design. 
As with real-world garments, 3D digital garments are represented by 2D patterns that specify how fabric pieces should be cut and assembled.
Designing such patterns today remains labor intensive, requiring expertise and specialized software.
This makes the automation of clothing design from images or text appealing; however, automated garment reconstruction 
remains largely unsolved.
A key challenge is the scarcity of paired sewing pattern–image data needed to train AI systems for this task.
While large-scale datasets of clothing images or 3D scans are increasingly available, acquiring image--pattern pairs remains extremely challenging. 
Importantly, creating such pairs requires expert knowledge and substantial manual annotation, which hinders large-scale data collection and makes data-driven approaches challenging to train, ultimately limiting their generalization.

To address this, Korosteleva and Sorkine-Hornung \cite{korosteleva2023garmentcode} introduced \textit{GarmentCode}, a domain-specific language (DSL) for sewing patterns that serves as a parametric garment configurator, and \textit{GarmentCodeData} \cite{korosteleva2024garmentcodedata}, a dataset generated by randomly sampling the configurator’s parameters. A garment can be represented with only $122$ GarmentCode parameters, which makes it a widely used representation for learning-based garment reconstruction methods 
that fine-tune vision language models (VLMs) to predict the garment parameters from images or structured text~\cite{bian2025chatgarment, zhou2025design2garmentcode}. 

Despite its convenience, GarmentCode representation is overparametrized: similar looking garments can be produced with a wide range of parameters, which makes it particularly challenging to learn precise mapping from images or text to GarmentCode. In addition, random combinations of parameters often yield unrealistic or inconsistent garments, leading to suboptimal generalization on in-the-wild images. For example, when assymmetric tops are sampled, the left and right parts may not form a plausible design (see Sup.Mat.).
Moreover, because the training data does not reflect realistic correlations between garment parameters (e.g. a t-shirt typically has both a crew neck and short sleeves), VLMs fine-tuned on randomly sampled parameters fail to learn such regularities. 
These limitations become more pronounced when garments are partially occluded (e.g. back side not visible, multi-layer outfits), which further restricts most existing methods to single-layer garment reconstruction.

At the same time, pre-trained VLMs already exhibit a strong ability to identify garment details from images, and prior methods have used this to label synthetic garments generated from GarmentCodeData~\cite{nakayama2025aipparel,bian2025chatgarment,zhou2025design2garmentcode}. Design2GarmentCode~\cite{zhou2025design2garmentcode} further demonstrates this potential by using a frozen multimodal understanding agent to answer structured questions about garments from images, sketches, or text. However, its pipeline still relies on fine-tuning a generation module that translates the extracted garment descriptions into valid GarmentCode design parameters. 
This suggests that the bottleneck is not garment understanding itself, but the gap between VLM-interpretable garment attributes and low-level sewing-pattern representations. More broadly, this points to a representation question: as VLMs become increasingly capable, should we rely on them to learn existing low-level representations, or should we design representations that better align with their natural descriptive abilities? While recent advances in large language models suggest the former \textit{when data is abundant}, garment modeling remains a low-data domain where an improved representation can make a significant difference.

Motivated by this, we propose \dsl{} (Natural Garment Language), a new DSL that expresses garment structure in descriptive, semantically meaningful terms that can be robustly queried via natural-language prompts. NGL represents garments using compact, human-interpretable attributes such as garment type, length, neckline shape, sleeve structure, fit, flare, and asymmetry, rather than requiring VLMs to predict low-level pattern parameters directly. We use NGL as the interface in a fully training-free inference pipeline, in which a pre-trained VLM extracts NGL specifications from an input image, and a deterministic parser converts them into GarmentCode parameters to generate valid sewing patterns. This representation eliminates the need for task-specific model training and naturally extends to multi-layer garments. 

We evaluate our method on the Dress4D \cite{wang20244d} and CloSe \cite{antic2024close} benchmarks, as well as on a newly collected set of approximately 250 in-the-wild fashion images. Our approach achieves state-of-the-art results on standard metrics such as Chamfer Distance and F-score, while also receiving higher perceptual scores in both human and GPT-based evaluations. 

In summary, our main contributions are:
\begin{itemize}
\item \dsl, a novel garment DSL optimized for VLM prompting, along with a deterministic parser that converts \dsl{} into GarmentCode parameters.
\item To the best of our knowledge, the first training-free approach for sewing pattern estimation from a single image, capable of handling both single-layer and multi-layer garments. %
\item An empirical demonstration that modern VLMs, when guided by domain knowledge, can match or surpass trained models in the garment reconstruction task.
\end{itemize}

\begin{figure*}
    \centering
    \includegraphics[width=\linewidth]{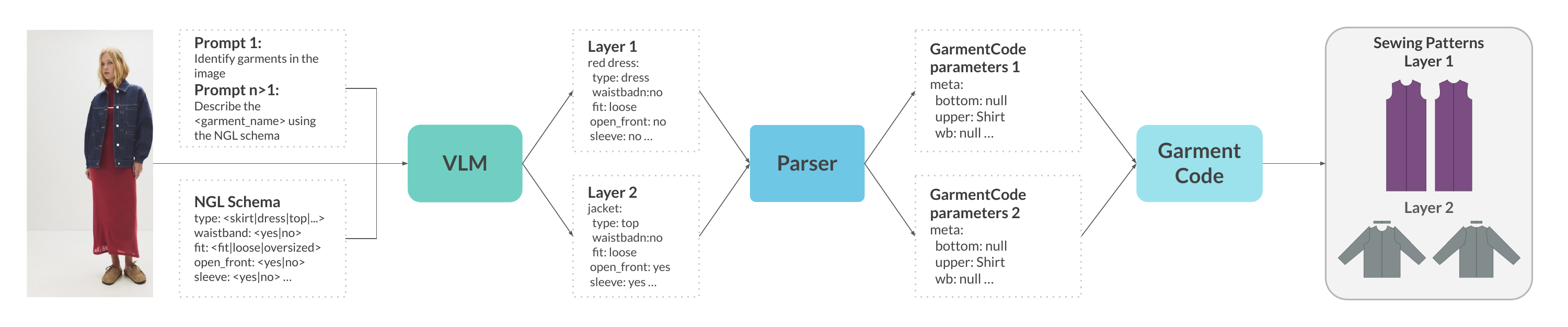}
    \caption{
        \qheading{Overview of the training-free pipeline with \moniker} Given an image containing a single- or multi-layer outfit, the method prompts a VLM in two stages: first, to identify garment types, then to extract attributes for each garment according to the NGL schema. A parser validates each attribute assigning a default for missing and invalid values, and then maps the results to GarmentCode design parameters. GarmentCode then generates the final sewing patterns.
    }
\label{fig:architecture}
\end{figure*}

\section{Related Work}
\label{sec:related}

\subsection{Garment Representations}
Prior work has explored several representations for 3D garment modeling. Broadly, these can be grouped into four different families:
\noindent \emph{(1) Explicit}, which directly model garment geometry using meshes~\cite{gundogdu19garnet}, point clouds~\cite{hong2021garment4D, Ma2022SkiRT}, or a canonical template that is subsequently deformed~\cite{zhao2023learning, chi2021garmentnets}. %
While these approaches can be effective and simple to optimize, they often rely on a fixed topology and offer a limited number of interpretable degrees of freedom. %
\noindent \emph{(2) Implicit}, which encode garments as continuous fields such as occupancy or signed distance functions (SDFs)~\cite{tiwari21neuralgif, Li2023ISP, corona2021smplicit, li2022dig, Moon2022ClothWild, aggarwal2022layered, dasgupta2025ngd, Santesteban2022ULNeF, xiu2022icon}, typically parameterized by neural networks. %
These representations are flexible and can capture fine geometric detail, but they usually require an explicit surface extraction step %
and do not naturally expose a semantic, edit-friendly structure. %
\noindent \emph{(3) 2D sewing panels}, where a garment is represented as a set of 2D panels together with stitching and assembly constraints that can be draped into 3D garments. Such representations are fabrication-aware and provide explicit control for editing by manipulating panel geometry and construction parameters. Some prior works learn a latent shape~\cite{Wang2018SharedShapeSpace} or a PCA-based model~\cite{Chen2022NeuralSewingMachine} over panel representation for garment design. Recent learning-based methods~\cite{liu2023sewformer, nakayama2025aipparel, li2025garmentdiffusion, Chen2024panelformer} have also explored inferring panels and stitches from images.
\noindent \emph{(4) Parametric approaches}, which represent garments using domain-specific languages (DSLs) from which  executable programs can generate 2D panel geometry and assembly 
\linebreak 
hierarchies~\cite{Chen2022NeuralSewingMachine, Pietroni2022CompPattern, korosteleva2021generating}. Currently, the most prominent and expressive parametric garment model is GarmentCode~\cite{korosteleva2023garmentcode}. 
Such structured, compositional representations are well suited for use with large language and vision–language models.

These representations differ in how much garment-specific domain knowledge they encode.
At the low end, meshes, point clouds, and implicit fields impose no garment specific priors, treating them as a generic 3D object. Sewing patterns sit in the middle, adopting panel and stich abstractions but leaving the geometry unconstrained. Parametric models such as GarmentCode go further, encoding garment construction regularities in a set of semantically meaningful parameters and aiming to guarantee validity by design. 
An important question is whether it remains productive to refine such representations through DSL design, %
or whether improved learning over existing representations  will eventually subsume these refinements.
When massive training datasets are available, end-to-end training of large models always proves effective.
However, sewing-pattern data is relatively scarce, and paired image–pattern data even more so.
We show that, in this situation, developing %
a structured intermediate representation is effective at mediating between images and patterns.

\subsection{Sewing Pattern Estimation from Images}
Prior work has addressed 2D sewing-pattern estimation from images using either optimization-based methods~\cite{Yang2016DetailedGR,Yang2018PhysicsInspiredGR} or learning-based approaches~\cite{liu2023sewformer, bian2025chatgarment, nakayama2025aipparel, zhou2025design2garmentcode}. 
Due to the difficulty of obtaining ground-truth paired image and sewing-pattern data, learning-based approaches often rely on synthetic data \cite{liu2023sewformer, he2024dresscode, Chen2024panelformer, Lim2024Spnet}.
Recent methods fine-tune large vision-language models (VLMs) on synthetic data randomly sampled from GarmentCode for multimodal sewing-pattern prediction.
AIpparel~\cite{nakayama2025aipparel} fine-tunes a large multimodal model (LLaVA \cite{liu2023llava}) using a tokenization scheme tailored to pattern representations with panel edges and stitches.
ChatGarment~\cite{bian2025chatgarment} fine-tunes LLaVA to predict GarmentCode parameters. %
The closest to our work is Design2GarmentCode, which uses a pre-trained VLM to infer a structured garment description that is then mapped to GarmentCode using a lightweight projector. The projector is trained on GarmentCodeData labeled with GPT-4V or %
through an inverse mapping from GarmentCode to textual descriptions.
While these training-based approaches show strong performance and functionality, they require curated supervision and remain limited by the randomly sampled nature of synthetic training data. This makes the learned mapping sensitive to the coverage and biases of the synthetic dataset, especially when applied to real-world images or free-form garment descriptions. 
In contrast, our approach is fully training-free and uses NGL as an explicit intermediate representation, allowing VLM-predicted garment attributes to be converted into GarmentCode parameters without learning %
from data.

\section{Method}
\label{sec:method}

Given a single-view image of a dressed person, our goal is to estimate sewing patterns for all garments present in the image. To this end, we introduce Natural Garment Language (NGL), a novel garment representation designed for inference with vision–language models (VLMs) and implemented as a semantic wrapper around GarmentCode (Section 3.2). We further propose a fully training-free inference pipeline that leverages the natural descriptive strengths of modern VLMs to predict NGL specifications from the image (Section 3.3). An overview of our method is shown in Figure 2.

\subsection{Background: GarmentCode}
\label{ssec:garmentcode}
GarmentCode~\cite{korosteleva2021generating} is a domain-specific language for parametric garment construction that assembles outfits from design parameters and body measurements. Design parameters are independent of body measurements and can therefore be used to generate sewing patterns for different body types. With a total of 122 parameters, GarmentCode is currently the most expressive parametric garment representation,  making it appealing for garment reconstruction methods.

The design parameters consist of a fixed set of high level template blocks (e.g. \textit{waistband}, \textit{shirt}, \textit{collar}) and a \textit{meta} block that selects the active templates. The \textit{meta} block contains a wide range of templates, including two bodice templates, two waistband templates, seven skirt templates and one pants template, many of which are subtypes of the other. For example, AsymmSkirtCircle extends SkirtCircle with asymmetric features, while SkirtLevels combines multiple skirt types. Moreover, the GarmentCode parameter space is not restricted to plausible designs, and some parameter combinations may produce invalid sewing patterns. 
These properties make GarmentCode expressive, but also difficult to learn from images or natural-language descriptions. This motivates our design of a compact, VLM-aligned DSL that sacrifices some of GarmentCode’s expressive power while exposing a more interpretable interface.

\begin{figure}
    \centering
    \includegraphics[width=1.0\linewidth]{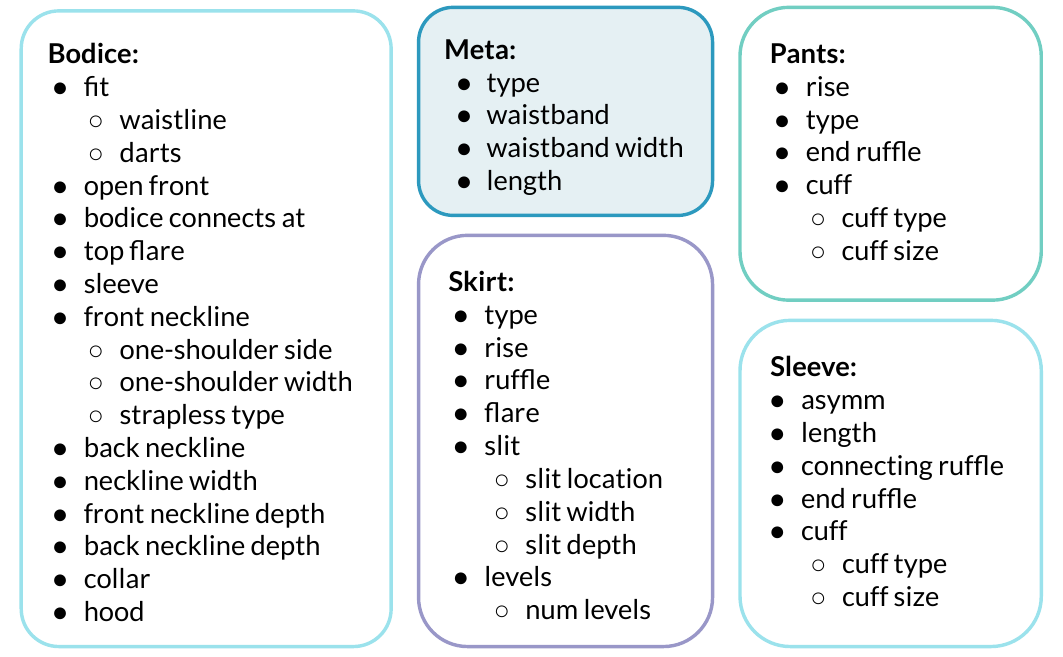}
    \caption{
        \qheading{\dsl{} parameter schema}
        We show NGL's five parameter blocks and the respective list of parameters. %
        \textit{Bodice} is activated when the type is dress, top or jumpsuit, \textit{sleeve} is activated then bodice sleeve parameter yields true, \textit{pants} are activated for pants and jumpsuit and \textit{skirt} is activated for skirt and dress.  Within the blocks, we show parameter hierarchy using bullet point levels.
    } \label{fig:ngl_all_params}
\end{figure}

\subsection{NGL}
\label{ssec:method_ngl}
We introduce \emph{Natural Garment Language (NGL)}, a novel domain-specific language specifically designed for training-free inference with large VLMs. As shown in \cref{fig:ngl_all_params}, the syntax consists of five parameter blocks: a \textit{meta} block used to set parameters common to all garments, 
and four blocks corresponding to major sewing pattern templates: \textit{bodice}, \textit{sleeve}, \textit{skirt} and \textit{pants}. 
The \textit{meta} block contains the \textit{type} parameter, which takes the values \textit{dress, top, skirt, pants} or \textit{jumpsuit} 
and determines the active template blocks, as well as parameters shared across all garment types (\textit{length}, \textit{waistband}, \textit{waistband\_width}). 
This design deliberately simplifies GarmentCode's original
template structure to reduce ambiguity and repetition resulting in only 44 parameters compared to 122 parameters in the original GarmentCode.

Each NGL parameter contains discrete value options, either boolean (yes/no) or natural language values. For example, the \texttt{length} parameter includes \textit{below the chestline}, \textit{mid-thigh}, and \textit{floor-length}, while \texttt{front\_neckline} and \texttt{back\_neckline} contain a predefined list of shapes to choose from (e.g.  \textit{crew | v-neck | one-shoulder | strapless etc.}).
To describe volumes of \texttt{flare} and \texttt{ruffle} we use adjectives such as \textit{low | medium | high}.
The parameters are interdependent and form a hierarchical structure, for instance, the \textit{sleeve} block is activated when the boolean parameter \texttt{bodice\_sleeve} is set true. %
The complete parameter schema can be found in \cref{fig:ngl_all_params}, and the full list of parameter options is available in Supplementary material.

To generate sewing patterns from NGL specifications, we use GarmentCode as an intermediate representation. GarmentCode is a suitable backbone because it already implements low-level sewing pattern construction and defines design parameters independently of body size, allowing the same garment specification to be instantiated on different bodies. The main challenge is that mapping NGL to GarmentCode is not one-to-one. GarmentCode provides a larger set of garment templates, each with its own parameterization; for example, its seven skirt templates are controlled by different parameter sets. In contrast, NGL uses a single compact skirt block with seven parameters that jointly cover nearly all skirt variations expressible in GarmentCode. To compile NGL into GarmentCode, our parser operates in two stages. First, it selects the corresponding GarmentCode template based on predefined rules; for example, the values in the NGL skirt block determine which GarmentCode skirt template is instantiated. Pseudocode for this template-selection procedure is provided in the supplementary material. Second, the parser applies template-specific conversion rules that map each NGL parameter to the corresponding GarmentCode parameters. This step is necessary because the same NGL parameter may be represented differently across GarmentCode templates. \cref{fig:length_dict} shows an example conversion rule for the NGL length attribute.

\begin{figure}
\centering
\begin{minted}[fontsize=\fontsize{6}{7.2}\selectfont]{python}
ShirtLengths = {
  'floor-length': 3.5,        'above-ankles': 3.3,
  'mid-calf': 3.05,           'slightly below the knee': 2.8,
  'knee-length': 2.55,        'slightly above the knee': 2.4,
  'mid-thigh': 2.2,           'short-mini': 2.0,
  'below the hips': 1.8,      'on the hips': 1.65,
  'above the hips': 1.5,      'below the waistline': 1.25,
  'on the waistline': 1.0,    'above the waistline': 0.9,
  'below the chestline': 0.7, 'on the chestline': 0.6,
}
SkirtLengths = {
  'floor-length': 0.8,        'above-ankles': 0.7,
  'mid-calf': 0.6,            'slightly below the knee': 0.5,
  'knee-length': 0.4,         'slightly above the knee': 0.3,
  'mid-thigh': 0.2,           'short-mini': 0.125,
  'below the hips': 0.1,      'on the hips': 0.0,
  'above the hips': -0.1,     'below the waistline': -0.2,
  'on the waistline': -0.25,  'above the waistline': -0.3,
  'below the chestline': -0.4,
}
PantsLengths = {
  'floor-length': 0.9,        'above-ankles': 0.8,
  'mid-calf': 0.65,           'slightly below the knee': 0.55,
  'knee-length': 0.45,        'slightly above the knee': 0.35,
  'mid-thigh': 0.25,          'short-mini': 0.15,
}
\end{minted}
\caption{\qheading{Example of mapping between \dsl{} \textit{length} parameter and the corresponding length in GarmentCode} We %
set GarmentCode values for each \dsl{} \textit{length} options for three GarmentCode template types.}
\label{fig:length_dict}
\end{figure}

\subsection{Inference pipeline}
\label{ssec:inference}
We implement a fully training-free pipeline that uses a  frozen VLM to infer NGL specifications from images and converts them into valid sewing patterns.

Given an input image, we first query the VLM to identify all visible garment layers and their ordering from inner to outer. Next, we process each garment layer individually: the VLM is given the input image,  and a single prompt with (1) the garment layer name,  %
(2) the NGL parameter schema with detailed descriptions of each parameter and value options, and (3) output instructions. The output is a structured NGL-aligned garment description. Precise prompts are available in supplementary material.
Finally, a parser validates the VLM's output by assigning a default for missing and invalid values ensuring successful pattern generation. The output is a structured NGL-aligned garment description, which can be converted into sewing patterns as described in \cref{ssec:method_ngl}.

Crucially, this pipeline does not require task-specific training or fine-tuning. As VLMs improve, the same pipeline can leverage stronger models to obtain better garment descriptions without changing the underlying system. This design demonstrates that a realistic sewing pattern can be obtained with state-of-the-art accuracy by aligning representations with what VLMs already know, rather than forcing models to learn low-level procedural detail.

\subsection{Textured Mesh Reconstruction}
\label{ssec:method_textures}

Optionally, we can use our approach to recover textured 3D meshes from a single image as follows. 
First, we estimate the human body shape and pose from the input image using TokenHMR \cite{dwivedi2024tokenhmr}, and align the GarmentCode body model with the estimated SMPL body, in order to obtain consistent body measurements. 
Given these measurements, we compile the predicted GarmentCode specifications into garment meshes corresponding to the estimated body size. We then repose individual garment meshes and assemble them into a single outfit using the method of Grigorev et al.~\cite{grigorev2024contourcraft}.

To recover garment appearance from in-the-wild images, we extract texture patches using a combination of Qwen2.5-VL-32B for garment localization and SAM \cite{kirillov2023segment} for segmentation. The extracted texture patches are normalized using FabricDiffusion \cite{zhang2024fabricdiffusion} to reduce illumination and appearance inconsistencies. The result is a fully textured 3D garment reconstruction that can be rendered and evaluated perceptually.

\section{Experiments}
\label{sec:experiments}
We evaluate \moniker{} on a combination of curated benchmarks and in-the-wild fashion images to assess both geometric reconstruction accuracy and generalization to real-world images. First, we quantitatively evaluate sewing pattern reconstruction on established datasets with ground-truth garment geometry (\cref{subsec:exp_img_recons}). Second, we conduct perceptual evaluations on diverse in-the-wild images, including challenging multi-layer outfits (\cref{ssec:results_perceptual}). Third, 
we perform ablations on the NGL parameter schema, level of detail, prompting strategy, and model size (\cref{sec:ablations}). 

\subsection{Baselines}
We compare against ChatGarment~\cite{bian2025chatgarment} and Design2\-GarmentCode~\cite{zhou2025design2garmentcode}, since both methods use parametric garment representation for reconstruction. 
We use two versions of ChatGarment: 
(i) the default version, which uses LLaVA~\cite{liu2023llava} as the base model and (ii) a GPT-powered version, which extracts coarse descriptions of each garment layer using GPT-5  and then uses both the description and the image as input to produce GarmentCode parameters, thus supporting multi-layer garment extraction. %
We refer to these as \emph{ChatGarment} %
and \emph{ChatGarment-GPT-5}, respectively.
Design2GarmentCode is originally powered by GPT-4V; however to ensure fair comparison between methods, we use GPT-5 for Design2GarmentCode as well. We run Design2GarmentCode up to three times on failed images to maximize the number of successfully processed examples.

\subsection{3D garment reconstruction}
\label{subsec:exp_img_recons}
To evaluate the geometric accuracy of the final outfit reconstruction, we follow the evaluation pipeline used in ChatGarment on the CloSe~\cite{antic2024close} and Dress4D~\cite{wang20244d} datasets. 
CloSe is a 3D clothing segmentation dataset containing 3,167 clothed-human scans with fine-grained clothing segmentation labels across 18 clothing categories, while 
Dress4D is a real-world 4D clothing dataset with 78k high-quality textured clothed-human scans. 
Dress4D covers 64 garment categories, including dresses, lower garments, upper garments, and outer garments.
We use the same image subsets as ChatGarment: 145 images from CloSe and 36 images from Dress4D with four loose fitting outfits. 
Following ChatGarment, we report two-way Chamfer Distance (CD) and F-score. In addition, we report the failure rate (FR), defined as the percentage of images for which a method does not produce a valid reconstruction.

\begin{table}[htbp]
  \centering
  \resizebox{\linewidth}{!}{
  \begin{tabular}{l |ccc | ccc}
    \toprule
    & \multicolumn{3}{c|}{Dress4D} & \multicolumn{3}{c}{CloSe}  \\
    \cline{2-7}
    Method & {CD ($\downarrow$)} & {F-Score ($\uparrow$)} & {FR ($\downarrow$)} & {CD ($\downarrow$)} & {F-Score ($\uparrow$)} & {FR ($\downarrow$)} \\
    \toprule
    ChatGarment & 3.99 & 0.78 & \SI{2.78}{\percent} & 3.59 & 0.76 & 0 \\
    ChatGarment* & 3.12 & 0.75 & 0 & 2.94 & \textbf{0.79} & 0 \\
    ChatGarment-GPT-5 & 5.41 & 0.78 & \SI{8.3}{\percent} & 8.45 & 0.73 & 0.68\si{\percent} \\
    Design2GarmentCode-GPT-5 & 2.69 & 0.79 & \SI{2.78}{\percent} & 2.93 & 0.76 & 1.38\si{\percent} \\
    \midrule
    \dsl-GPT-5 & \textbf{1.75} & \textbf{0.81} & 0 & \textbf{2.79} & 0.78 & 0.68\si{\percent} \\
    \bottomrule
  \end{tabular}
  }
  \caption{\textbf{Quantitative evaluation on the Dress4D and CloSe datasets}. 'ChatGarment' reports results from our re-evaluation, and 'ChatGarment*' reports values from the original paper. FR refers to failure rate. }
    \label{table:quant_dress4d_close}
\end{table}

Results are shown in ~\cref{table:quant_dress4d_close}. %
Our method consistently outperforms the baselines on both datasets without any fine-tuning, with a stronger improvement on the more challenging Dress4D subset containing loose-fitting clothing. Failure cases are rare and typically correspond to invalid outputs or model refusal on a small number of images. We include FR for completeness and compute the remaining metrics on successfully generated reconstructions.
Qualitative results are shown in \cref{fig:figures_dress4d_close}.

\subsection{Perceptual evaluation}
\label{ssec:results_perceptual}

Since datasets with ground truth image-garment mesh pairs, such as Dress4D and CloSe, are limited in style and diversity of outfits, we manually select 154 single-layer and 99 multi-layer images spanning diverse styles from the fashion website \href{https://www.asos.com}{asos.com}.

\begin{table}
\resizebox{\linewidth}{!}{
\centering %
\begin{tabular}{l|c c|c c}
\toprule
\multirow{2}{*}{Method} & \multicolumn{2}{c|}{Single-layer (146 images)} & %
    \multicolumn{2}{c}{Multi-layer (98 images)}\\
\cline{2-5}
\multirow{2}{*}{}
 & Mean ($\uparrow$) & FR ($\downarrow$)
 & Mean ($\uparrow$) & FR ($\downarrow$)\\
\cline{1-5}
ChatGarment
    & 2.678 $\pm$ 0.286 & 0
    & 1.633 $\pm$ 0.133 & \SI{1}{\percent}\\
ChatGarment-GPT-5
    & 2.842 $\pm$ 0.233 & 0
    & 2.061 $\pm$ 0.221 & 0 \\
Design2GarmentCode-GPT-5
    & 4.438 $\pm$ 0.258 & \SI{5.2}{\percent}
    & 1.806 $\pm$ 0.181 & 0 \\
\midrule
NGL-GPT-5
    & \textbf{5.692 $\pm$ 0.253} & 0
    & \textbf{5.163 $\pm$ 0.398} & 0 \\   
\bottomrule

\end{tabular}
}
\caption{\textbf{AI study on single-layer and multi-layer images from the ASOS dataset.} 
Mean is reported with 95\% confidence intervals (CI).
The scores take into account only the images that were successfully processed by each method.  \moniker{} particularly outperform others on multi-layer images. FR refers to failure rate.} 
\label{table:asos}
\end{table}

We conduct two perceptual studies: an AI-based study and a human perceptual study. 
For the \emph{AI study}, we prompt GPT-5.0 to rate textured renderings of the reconstruction on a 0-9 scale, where 0 corresponds to an entirely incorrect garment type and 9 corresponds to a highly accurate reconstruction that captures both the overall garment structure and finer details such as cuffs, frills, and hems. The full prompt is provided in the supplementary material. 
For the \emph{human perceptual study}, participants compare outfit renderings from two methods and indicate which sewing-pattern reconstruction better matches the original image on a scale from -2 to 2, where positive values indicate a preference for the \dsl-based method and negative values indicate a preference for the baseline.

For the human study, we compare \dsl-GPT-5 against~Design2\-Garment\-Code-GPT-5 on single-layer outfits, and against Chat\-Garment-GPT-5 on multi-layer outfits, since ChatGarment-GPT-5 is the only baseline that supports multi-layer reconstruction. The results, shown in \cref{table:human_study}, 
are consistent with the AI-based study: participants strongly prefer NGL-GPT-5 over ChatGarment-GPT-5 on multi-layer images with a mean preference score of $1.44 \pm 0.08$ (95\% confidence interval), and also prefer NGL-GPT-5 over Design2GarmentCode-GPT-5 on single-layer images with a mean preference score of $0.56 \pm 0.15$ (95\% confidence interval), indicating a moderate but reliable preference. 
Additional details on the suman study are provided in the supplementary material.

\begin{table}
\resizebox{\linewidth}{!}{
\centering %
\begin{tabular}{llcccc}
\toprule
NGL-GPT-5 versus: & Outfit type & N images & Mean ($\uparrow$) \\
\midrule
 ChatGarment-GPT-5 & Multi-layer & 97 & $1.44 \pm 0.08$  \\
Design2GarmentCode-GPT-5 & Single-layer & 143 & $0.56 \pm 0.15$  \\
\bottomrule
\end{tabular}
}
\caption{\textbf{Human study on single-layer and multi-layer garment images from the ASOS dataset}. Positive values indicate preference for \dsl{}-GPT-5. On average, 8 participants successfully completed the study. Mean preference scores are averaged per image and reported with 95\% confidence intervals (CI).}
\label{table:human_study}
\end{table}

\subsection{Ablations}
\label{sec:ablations}

We conduct ablations to analyze three aspects of NGL: (i) which NGL parameters VLMs can reliably infer from images; (ii) how simplifying the parameter schema affects reconstruction quality; and (iii) how prompting strategy affects performance. Throughout these experiments, we evaluate models of different scales, including GPT-5.0 \cite{openai_chatgpt}, Qwen-3-VL-Instruct \cite{qwen3} with 235B and 30B parameters, and Qwen-2.5-VL-Instruct \cite{qwen25-VL} with 72B, 32B, 7B, and 3B parameters.

\subsubsection{\dsl{} parameter schema ablation}
\label{ssec:llm_garment_knowledge}
Given the \dsl{} schema with 44 parameters, we conduct an experiment to assess whether VLMs can recover these parameters from images. 
To this end, we construct a labeled dataset by manually selecting real garment images for each parameter value (e.g., different neckline types), aiming to include at least two examples for uncommon values (e.g., one sleeve, side slit) and more than five examples for common values.
This results in a dataset of 164 partially labeled images, which we denote as \asoslabeled{} and use to evaluate the zero-shot performance of VLMs.
We report F1-score as the evaluation metric to account for class imbalance across parameter values, such as one-sleeve tops being less common than two-sleeve tops.
\begin{figure}
  \includegraphics[width=0.5\textwidth]{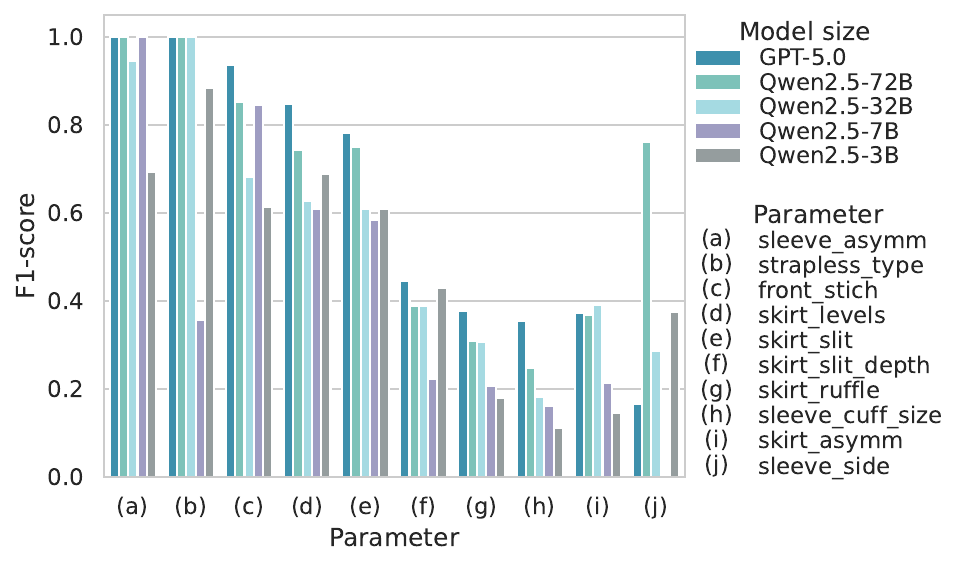}
    \caption{\qheading{Empirical results on VLMs knowledge about garments} The plot shows the F1 score computed on \asoslabeled{} dataset across selected set of parameters from \dsl{} and various model sizes. All models can confidently identify intricate garment details that are commonly described on fashion websites (e.g. \textit{strapless\_type} straight or heart-shaped strapless neckline), but struggle with details that are not commonly described (e.g. \textit{skirt\_asymm} skirt with the back part longer than the front or one side longer than the other)}
  \label{fig:good_bad_params}
\end{figure}

Our analysis, summarized in Fig.~\ref{fig:good_bad_params}, 
shows that large VLMs can identify some non-trivial garment details, but not all of them. 
For example, models can detect the presence of a skirt slit, but struggle to estimate its precise depth. 
We find that current state-of-the-art VLMs are strongest at recognizing attributes commonly used in fashion descriptions, while they struggle with details that are less commonly described, such as precise ruffle volume or cuff size.

We also observe that GPT-5 can identify intricate garment details such as cuff type or skirt-slit location, whereas smaller models perform significantly worse. We therefore introduce a simplified schema to evaluate whether zero-shot garment reconstruction remains possible with smaller models. We remove non-essential components such as cuffs, skirt slits, skirt levels, sleeve asymmetry, bodice darts, and bodice flare. We also simplify flare and ruffle parameters to binary values and reduce the number of value options for length, fit, and rise. We refer to this simplified version as \dsl-0 and the full version as \dsl-1.

\begin{figure}
  \includegraphics[width=0.48\textwidth]{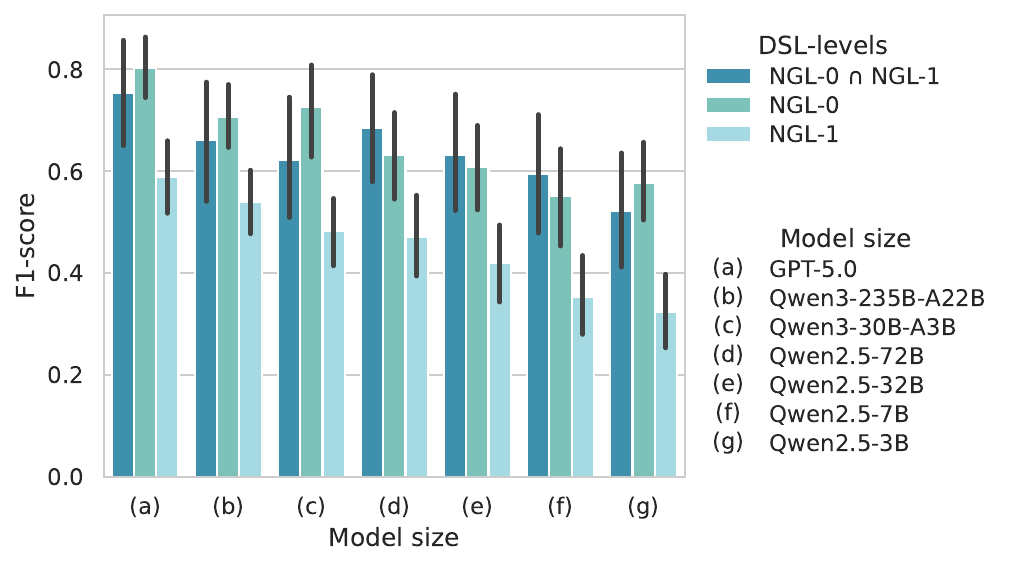}
  \caption{\qheading{Quantitative results on Garment Attribute Accuracy across \dsl{} LODs} We report the F1 score on our \asoslabeled{} dataset to evaluate the prediction quality accross \dsl{} parameters and model sizes. \dsl{}-0 $\cap$ \dsl{}-1 denotes the subset of attributes shared by both schema versions.}
  \label{fig:lod_f1}
\end{figure}

We compute the mean F1-score over parameters shared between \dsl-0 and \dsl-1 (\dsl-0 $\cap$ \dsl-1), and parameters unique to \dsl-0 and \dsl-1.
Results are shown in~\cref{fig:lod_f1}. 
We observe that most models accurately infer the basic garment attributes included in \dsl-0, such as garment type, sleeve presence, and open front. As expected, performance improves with model size, with larger VLMs consistently producing more accurate attribute predictions. Performance decreases for the more complex stylistic details included in \dsl-1; however, larger models still achieve reasonable score under this setting.

\subsubsection{Level of Detail (LOD) and prompting strategy ablation}
\label{ssec:lod_ablation}

As described above, we define two levels of detail: \dsl-0 and \dsl-1. We compare them on the image based reconstruction task using the AI-based evaluation protocol describe in \cref{ssec:results_perceptual}.
In addition, we evaluate two prompting strategies. 
In \textit{single-shot prompting}, the model is queried with all NGL parameters at once, as described in \cref{ssec:inference}. In \textit{sequential prompting}, the model is queried one parameter at a time, with each question selected based on previous responses following the \dsl{} hierarchical schema.

\begin{figure}
  \includegraphics[width=0.48\textwidth]{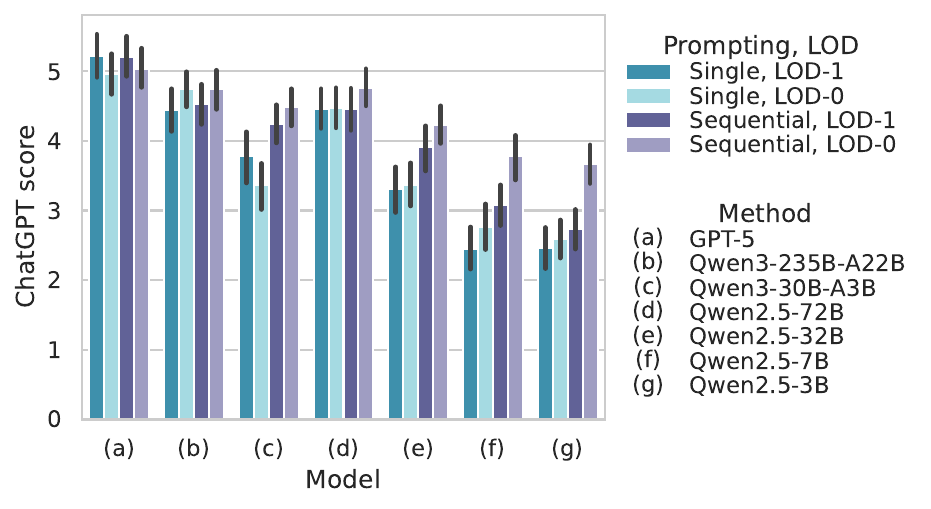}
  \caption{\qheading{Prompting strategy and level of detail ablation} Only GPT-5 is able to benefit the additional details present in \dsl{}-1. Smaller models are able to achieve reasonable reconstruction with \dsl{}-0 under sequential prompting strategy.}
  \label{fig:abl_lod_prompt}
\end{figure}

\begin{figure*}
\centering
  \includegraphics[width=0.93\textwidth]{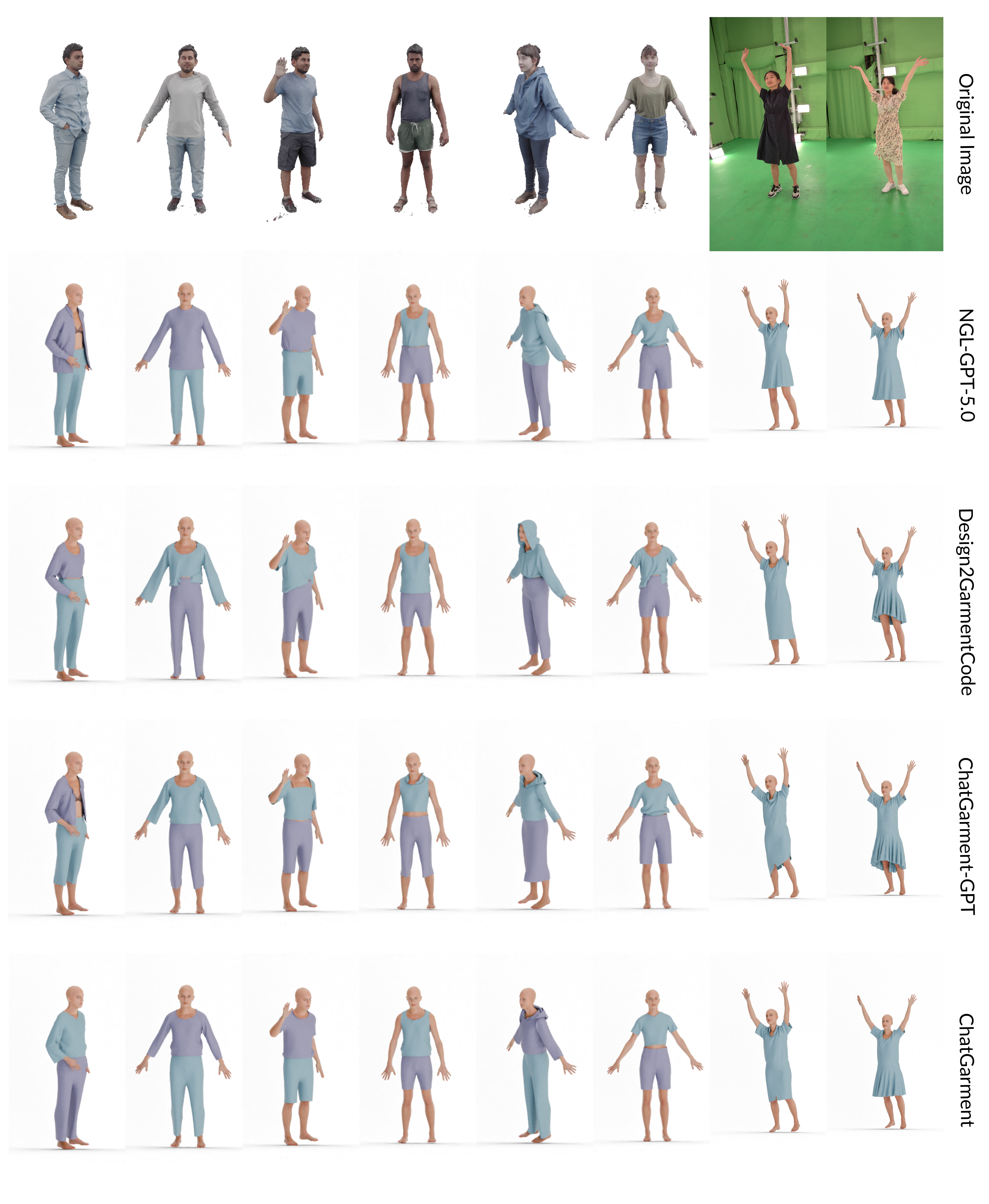}
  \caption{\textbf{Qualitative comparisons on the CloSe and Dress4D datasets}. We compare our method (\dsl{}-GPT-5.0) with Design2GarmentCode, ChatGarment-GPT, and ChatGarment. Columns 1-6 show results from the CloSe dataset, while columns 7-8 show from the Dress4D dataset. \dsl{} correctly captures garment shapes and lengths, which results in a lower Chamfer Distance to the original mesh. 
  }
  \label{fig:figures_dress4d_close}
\end{figure*}

Results in \cref{fig:abl_lod_prompt} show that the frontier model, GPT-5.0, performs similarly well with \textit{single shot prompting}  and \textit{sequential prompting}. Moreover, GPT-5.0 is the only model that reliably benefits from the additional garment details included in 
\dsl-1. In contrast, smaller models perform better 
 with the simplified \dsl{}-0 schema under sequential prompting.

\section{Conclusion}

We present NGL, a domain-specific language that enables fully training-free sewing-pattern estimation with VLMs. Our approach is motivated by the observation that modern vision–language models possess strong garment knowledge, yet struggle to learn low-level sewing-pattern representations. Rather than relying on additional training data, we design a representation that aligns with the natural descriptive abilities of VLMs. This design enables accurate sewing-pattern reconstruction without model fine-tuning and naturally extends beyond the single-layer setting to support multi-layer outfit reconstruction under occlusion. More broadly, our results demonstrate that representation design can effectively address data scarcity in garment reconstruction.

As with any parametric garment representation, our method is limited by its predefined parameter space and cannot reconstruct sewing patterns outside this space. Nevertheless, templates are widely used in garment construction, and expanding the variety of templates and their possible combinations could further improve the quality and expressiveness of reconstructed sewing patterns.

\FloatBarrier
\bibliographystyle{configs/ieeenat_fullname}
\bibliography{main}

\begin{figure*}
\centering
  \includegraphics[width=0.93\textwidth]{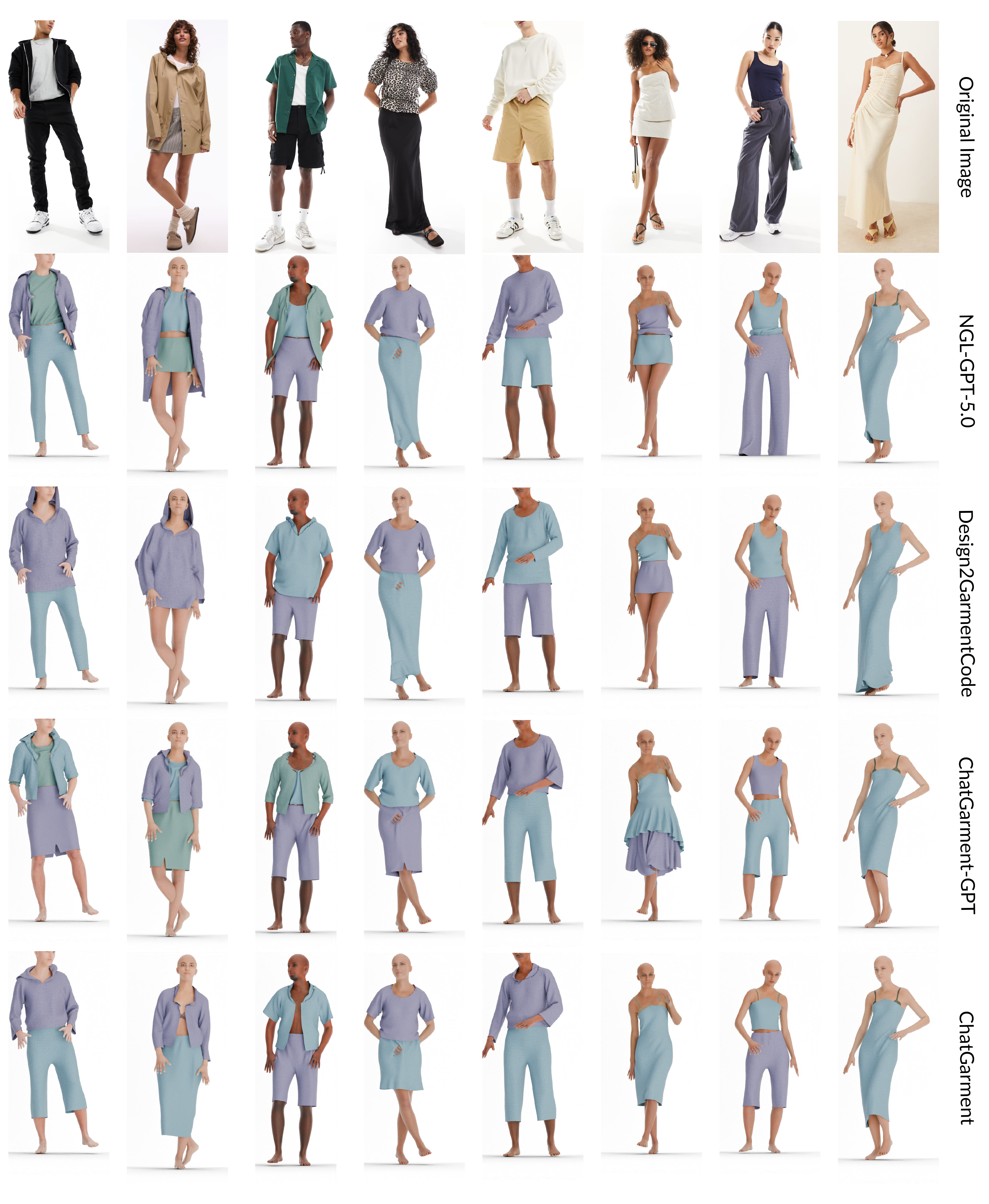}
  \caption{\textbf{Qualitative results from the \asosdatasetname{} dataset}.  We compare our method (\dsl{}-GPT-5.0) with Design2GarmentCode, ChatGarment-GPT, and ChatGarment. \moniker{} accurately captures garment details on both multi-layer garments (columns 1-3) and single-layer garments (columns 4-8), particularly outperforming others on details such as neckline shapes.
  }
  \label{fig:figures_single_multi}
\end{figure*}

\end{document}